\documentclass[a4paper, 12pt, english]{article}

\usepackage[utf8]{inputenc}
\usepackage{amsmath,amssymb}
\usepackage{graphicx}
\usepackage{subfig}
\usepackage[colorinlistoftodos]{todonotes}
\usepackage[natbibapa]{apacite}
\usepackage{enumitem}
\usepackage{indentfirst}
\usepackage{verbatim}
\usepackage{textcomp}
\usepackage{gensymb}
\usepackage{titlesec}
\usepackage{microtype}
\usepackage{xurl}
\usepackage{makecell}
\usepackage{relsize}
\usepackage{multirow} 
\usepackage{abstract}

\usepackage{CJKutf8}

\usepackage{lipsum}
\usepackage{xcolor}
\usepackage{xparse}
\NewDocumentCommand{\myrule}{O{1pt} O{2pt} O{black}}{%
  \par\nobreak 
  \kern\the\prevdepth 
  \kern#2 
  {\color{#3}\hrule height #1 width\hsize} 
  \kern#2 
  \nointerlineskip 
}
\usepackage[section]{placeins}

\usepackage{booktabs}
\usepackage{colortbl}%

\usepackage{etoolbox}

\usepackage[colorlinks,citecolor=black,urlcolor=blue,bookmarks=false,hypertexnames=true,citecolor=blue]{hyperref} 
\usepackage[capitalize,nameinlink]{cleveref}

\usepackage{geometry}
\begin{document}


\makeatletter
\renewcommand{\@maketitle}{
  \newpage
  \null
  \vskip 2em 
  \begin{center}
  {\LARGE \bfseries \@title \par} 
  \vskip 1.5em 
  {\large 
   \lineskip .5em 
   \@author \par}
  \vskip 1em 
  \end{center}
  \par
  \vskip 1.5em 
}
\makeatother

\author{
  \normalfont \textbf{Step-Audio Team} \\ 
  StepFun \\ 
}

\title{Step-Audio: Unified Understanding and Generation in Intelligent Speech Interaction}

\maketitle

\begin{abstract}
\noindent Real-time speech interaction, serving as a fundamental interface for human-machine collaboration, holds immense potential. However, current open-source models face limitations such as high costs in voice data collection, weakness in dynamic control, and limited intelligence. To address these challenges, this paper introduces \textbf{Step-Audio}, the first production-ready open-source solution. Key contributions include: 1) a 130B-parameter unified speech-text multi-modal model that achieves unified understanding and generation, with the Step-Audio-Chat version open-sourced; 2) a generative speech data engine that establishes an affordable voice cloning framework and produces the open-sourced lightweight Step-Audio-TTS-3B model through distillation; 3) an instruction-driven fine control system enabling dynamic adjustments across dialects, emotions, singing, and RAP; 4) an enhanced cognitive architecture augmented with tool calling and role-playing abilities to manage complex tasks effectively. Based on our new \textbf{StepEval-Audio-360} evaluation benchmark, Step-Audio achieves state-of-the-art performance in human evaluations, especially in terms of instruction following. On open-source benchmarks like LLaMA Question, shows \textbf{9.3\%} average performance improvement, demonstrating our commitment to advancing the development of open-source multi-modal language technologies. Our code and models are available at \url{https://github.com/stepfun-ai/Step-Audio}. 
\end{abstract}










\section{Introduction} 
\label{sec:intro}
\noindent The evolution of artificial intelligence toward general-purpose systems has positioned real-time speech interaction as a critical interface for human-machine collaboration. While recent multi-modal large language models (LLMs) have accelerated progress in this domain, open-source communities face persistent challenges despite breakthroughs in proprietary systems like GPT-4o~\citep{hurst2024gpt} and Doubao~\citep{doubaovoice}. Existing open-source models such as Qwen2-Audio~\citep{chu2024qwen2audiotechnicalreport}, Llama 3~\citep{dubey2024llama} and wavLLM~\citep{hu2024wavllm} struggle with three fundamental limitations: the separation of understanding and generation processes that impedes end-to-end system integration, dependence on laborious manual speech data acquisition methods that restricts efficient voice replication, and inadequate precision in regulating prosodic features, regional dialects, and tool utilization capabilities. These limitations highlight the urgent demand for deployable frameworks that harmonize streamlined architecture with dual competencies in affective computing (accurate emotion perception and adjustment) and contextual cognition (situational reasoning and response formulation).

\noindent Current open-source speech systems confront multiple architectural challenges. 
The traditional framework employs a cascading approach~\citep{huang2023audiogptunderstandinggeneratingspeech} combining Automatic Speech Recognition
(ASR), LLM processing, and Text-to-Speech (TTS). This framework introduces error propagation through modality transitions while increasing system complexity. Pure end-to-end approaches, though conceptually elegant, often sacrifice performance in open-domain dialogue quality~\citep{zeng2024glm4voiceintelligenthumanlikeendtoend}. The tension between modular design and fully integrated systems remains unresolved. Furthermore, traditional text-to-speech pipelines depend on manually curated datasets, particularly for multilingual and multidialect scenarios—a process requiring prohibitive human annotation effort. Existing solutions also lack sophisticated control mechanisms for dynamic speech adaptation, such as real-time adjustment of speaking rate, emotional prosody, or musical rendering (e.g., Singing and RAP vocals). Crucially, the absence of tool invocation capabilities and contextual awareness prevents handling complex queries like ``Retrieve live weather data and report it in Cantonese," necessitating manual API integration.

\noindent This report presents Step-Audio, the first production-ready open-source framework for intelligent speech interaction that harmonizes comprehension and generation through four key innovations. 

\begin{itemize}
    \item \textbf{130B-Parameter Multi-modal Model}: A single unified model integrating comprehension and generation capabilities, performing speech recognition, semantic understanding, dialogue, voice cloning, audio editing and speech synthesis. We have made the 130B Step-Audio-Chat variant open source.
    
    \item \textbf{Generative Data Engine}: Eliminates traditional TTS's reliance on manual data collection by generating high-quality audio through 
    our 130B-parameter multi-modal model. Leverages this data to train and publicly release a resource-efficient Step-Audio-TTS-3B model with enhanced instruction-following capabilities for controllable speech synthesis.
    
    \item \textbf{Granular Voice Control}: Enables precise regulation through instruction-based control design, supporting multiple emotions (anger, joy, sadness), dialects (Cantonese, Sichuanese, etc.), and vocal styles (RAP/Singing, a cappella humming) to meet diverse speech generation needs.
    
    \item \textbf{Enhanced Intelligence}: Improves agent performance in complex tasks through ToolCall mechanism integration and role-playing enhancements.
\end{itemize}

\begin{figure}[!ht]
    \centering
    \includegraphics[width=0.85\textwidth]
    {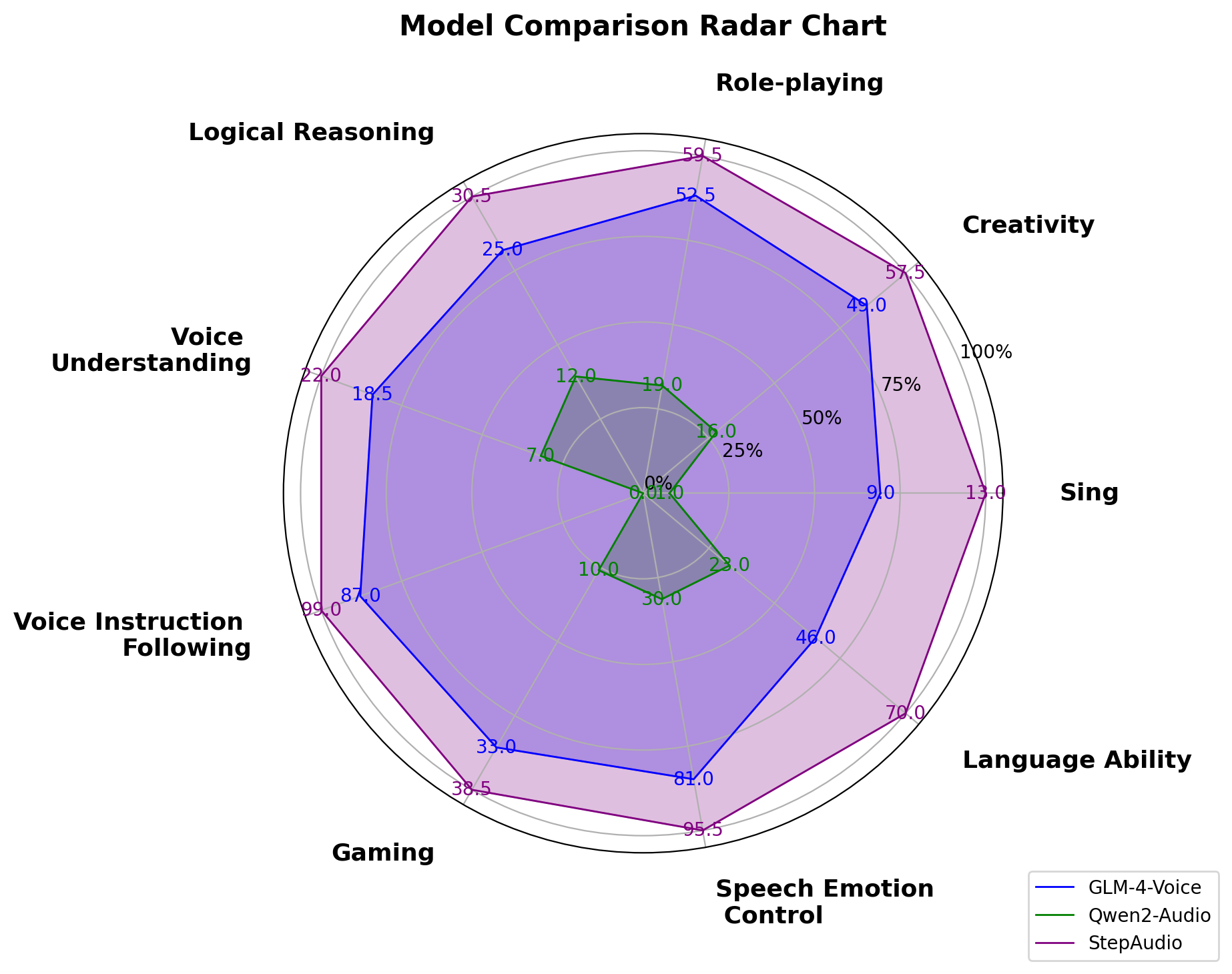}
    \caption{
    \textbf{Human Evaluation of End-to-End Speech Interactions.} We conduct comprehensive human assessments comparing Step-Audio against GLM-4-Voice~\citep{zeng2024glm4voiceintelligenthumanlikeendtoend} and Qwen2-Audio~\citep{chu2024qwen2} across nine critical dimensions: role-playing, logical reasoning, creativity, singing language ability, speech emotion control, gaming interaction, voice instruction following, and voice understanding. Expert evaluators rated end-to-end dialog sessions using Likert scales (1-5) for naturalness and task completion. Step-Audio represents the state-of-the-art (SoTA) across all these dimensions. It is particularly remarkable in language ability, demonstrating a high level of proficiency in grammar, semantics, and language generation. In singing, Step-Audio outshines the other models with its natural pitch control, rhythm accuracy, and overall harmonious vocal output, making it a top - tier performer in these two crucial aspects.}
    \label{fig:huamn_eval}
\end{figure}

\noindent In open-source benchmarks, Step-Audio demonstrates exceptional performance. It achieves SoTA results on open-domain question answering and complex instruction tasks including LLaMA Question, TrivialQA, and ComplexBench, with an average improvement of 9.3 points compared to the best open-source metrics, validating its advantage in generalized deep semantic understanding capabilities. Additionally, to address the current lack of comprehensive end-to-end speech dialogue evaluation systems, we introduce the multi-dimensional StepEval-Audio-360 evaluation framework covering 9 dimensions, including logical reasoning, creative ability, language proficiency, and comprehension control among other key capabilities. As shown in Figure \ref{fig:huamn_eval}, Step-Audio achieves SoTA results across all dimensions in subjective comparisons against open-source models like GLM-4-Voice and Qwen2-Audio, with improvements of 19.2\%, 23.7\%, and 43.2\% in response quality, response relevance, and factual accuracy respectively. Particularly in generation control dimensions such as emotion understanding, speech rate control, RAP vocals, and role-playing, compared to open-source SoTA models, the IF (Instruction Following) and MOS (Mean Opinion Score) metrics improved by 29.8\% and 27.1\% respectively, highlighting its leading advantage in complex speech interaction scenarios.

\section{Related Work}
\noindent Recent progress in end-to-end speech systems have markedly improved human-AI audio interaction. Early approaches relied on cascaded ASR-LLM-TTS pipelines~\citep{huang2023audiogptunderstandinggeneratingspeech}, where distinct modules for speech recognition, language modeling, and speech synthesis are sequentially connected. However, these systems suffered from latency buildup, error propagation, and disjointed optimization. Later approaches sought to enhance integration by directly linking speech encoders to LLMs through trainable adapters~\citep{kong2020pannslargescalepretrainedaudio, chu2024qwen2audiotechnicalreport, das2024speechverselargescalegeneralizableaudio}, though they still required separate TTS modules for audio output. 

\bigskip
\noindent The emergence of fully end-to-end systems marked a paradigm shift. Architectures like Llama-Omni~\citep{fang2024llamaomniseamlessspeechinteraction} integrated non-autoregressive~(NAR) TTS modules with language models, using connectionist temporal classification (CTC) loss. Freeze-Omni~\citep{wang2024freezeomnismartlowlatency} uses a combination of autoregressive and NAR speech decoders. These systems demonstrated improved latency but exhibited limitations in handling emotional nuance and natural conversational flow. MinMo~\citep{chen2025minmomultimodallargelanguage} introduced autoregressive speech token prediction through the CosyVoice2~\citep{du2024cosyvoice2scalablestreaming} decoder, while interleaved modeling approaches~\citep{zeng2024glm4voiceintelligenthumanlikeendtoend, nguyen2024spiritlminterleavedspoken} alternated between text and speech token generation at the sequence level.

\bigskip
\noindent Parallel decoding architectures like Moshi~\citep{2024moshispeechtextfoundationmodel} and Mini-Omni~\citep{xie2024miniomnilanguagemodelshear} represented a significant leap by generating text and multiple speech codebook tokens simultaneously. These systems achieved lower latency through compressed speech token sequences but faced challenges in preserving linguistic capabilities when scaling speech token bandwidth. Current systems generally specialized in specific aspects: GLM-4-Voice~\citep{zeng2024glm4voiceintelligenthumanlikeendtoend} prioritized latency reduction, while Moshi emphasized speech quality, but none holistically addressed emotion awareness, conversational naturalness, and real-time knowledge integration.

\bigskip
\noindent Recent methodological advances have systematically investigated emotion-aware interaction paradigms, though their integration with multi-modal frameworks remains nascent. While some systems~\citep{wang2024freezeomnismartlowlatency} incorporated basic sentiment analysis, they lacked bidirectional emotional resonance-neither detecting paralinguistic cues in user speech nor generating contextually appropriate emotional responses. The naturalness gap persisted due to LLMs' tendency toward verbose, text-optimized outputs~\citep{fang2024llamaomniseamlessspeechinteraction}, ill-suited for spoken dialogue. Recent work has introduced task-specific optimizations: LUCY~\citep{gao2025lucylinguisticunderstandingcontrol} adopted the architectural framework of Mini-Omni~\citep{xie2024miniomnilanguagemodelshear}, augmented with specialized fine-tuning on conversational datasets for emotion control and function-calling.


\section{Architecture}
\begin{figure}[!ht]
 			\begin{center}
				\includegraphics[width=145mm]{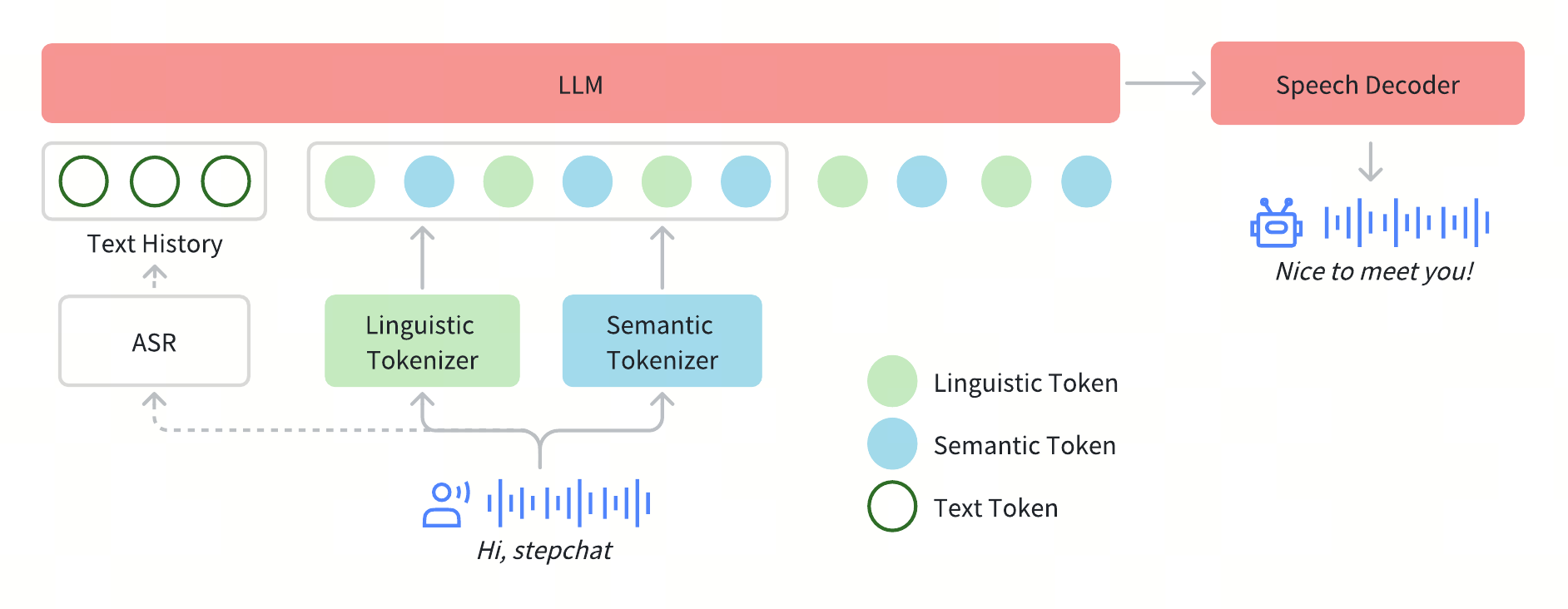}
			\end{center}
       		\caption{\label{fig:Highpass} Architecture of Step-Audio. Step-Audio primarily consists of three components: the speech tokenizer, the LLM, and the speech decoder. The speech tokenizer is responsible for discretizing the input speech into tokens. The LLM models both text and speech tokens, while the speech decoder generates the waveform output.}
\end{figure} 
\noindent Traditional voice dialogue systems typically employ a cascaded architecture comprising ASR, LLM, and TTS modules. However, our proposed model, having undergone comprehensive multi-modal training and alignment of text and audio during the pretraining phase, already possesses end-to-end voice dialogue capabilities. Despite extensive exploration of alternative designs, we ultimately adopted the AQTA (audio input, text output) + TTS framework for real-time voice dialogue as shown in Figure \ref{fig:Highpass}, driven by the following considerations:

\begin{itemize}
    \item \textbf{Scarcity of high-quality pure-voice dialogue data}: The limited availability of pure-voice dialogue data, coupled with its constrained scenarios, restricts the training efficiency of end-to-end voice dialogue models.
    
    \item \textbf{Controllability and customization of output speech}: By incorporating a TTS module, we gain flexible control over speech parameters such as timbre and pitch to meet users' personalized demands, while continuously enhancing the model's expressive capabilities.
\end{itemize}
Our goal is to establish Step-Audio as a real-time multi-modal model that seamlessly integrates speech understanding and synthesis through four key components: (1) A dual-codebook tokenization framework employing parallel linguistic (16.7Hz, 1024-codebook) and semantic (25Hz, 4096-codebook) tokenizers with 2:3 temporal interleaving; (2) A 130B-parameter LLM based on Step-1~\citep{step1}, enhanced through audio-contextualized continual pretraining and postraining; (3) A hybrid speech synthesizer combining with flow matching and neural vocoder, optimized for real-time waveform generation. In addition, a Voice Activity Detection (VAD) module was employed to extract vocal segments.

\subsection{Tokenizer}
\noindent To overcome the limitations of conventional speech tokenizers, which separately capture information for understanding or generation task, we propose a dual-codebook speech tokenizer framework in Step-Audio similar to ARCON~\citep{ming2024advancingautoregressivecontinuationvideo}. This approach employs two distinct tokenizers, linguistic and semantic, to better represent speech features. The linguistic tokenizer is utilized to extract structured, high-level representations, including phonemic and linguistic features, whereas the semantic tokenizer is designed to encode both semantic and coarse-grained acoustic characteristics.

\bigskip

\noindent For linguistic tokenization, we utilize the output from the Paraformer~\citep{gao2023paraformerfastaccurateparallel} encoder, which is quantized into discrete representations at a token rate of 16.7 Hz. For semantic tokenization, we employ CosyVoice's~\citep{du2024cosyvoicescalablemultilingualzeroshot} tokenizer, specifically designed to efficiently encode features essential for generating natural and expressive speech outputs, operating at a token rate of 25 Hz. The linguistic tokenizer employs a codebook size of 1024, while the semantic tokenizer utilizes a larger codebook size of 4096 to capture finer acoustic details. 

\bigskip
\noindent To effectively integrate these two tokenization schemes, we implement a token-level interleaving approach inspired by SpiritLM~\citep{nguyen2024spiritlminterleavedspoken}. Given the differing token rates, we establish a temporal alignment ratio of 2:3, where every two linguistic tokens are paired with three semantic tokens.

\subsection{LLM}
\noindent To enhance Step-Audio's ability to effectively process speech information and achieve accurate speech-text alignment, we conducted audio continual pretraining based on Step-1, a 130-billion parameter pretrained text-based LLM. The details of the pretrain and post-train processes for Step-Audio are comprehensively discussed in section \ref{chap:pretrain} and \ref{chap:posttrain}.

\bigskip
\noindent In multi-turn dialogue systems, the substantial disparity in length between audio tokens and text tokens necessitates efficient processing strategies. To address this, historical information is initially transcribed into textual format utilizing an ASR model prior to system input, thereby optimizing computational efficiency. However, it should be noted that the model architecture maintains the capability to process and utilize audio tokens as historical context when required.


\subsection{Speech Decoder}
\noindent Speech decoder consists of a 3-billion parameter language model, a flow-matching model and a mel-to-wave vocoder primarily designed to receive text or audio tokens and generate continuous time-domain stylized waveform that incorporate historical information and instructions. To optimize the intelligibility and naturalness of the synthesized speech, the speech decoder is trained using a dual-code interleaving approach, ensuring seamless integration of linguistic and semantic features throughout the generation process. On a speech decoder with a larger parameter, we have observed the emergence of enhanced generative capabilities. For further details, please refer to section \ref{sec:tts}. 

\subsection{Real-time Inference}
\begin{figure}[htbp]
  \centering
  \includegraphics[width=1.0\textwidth]{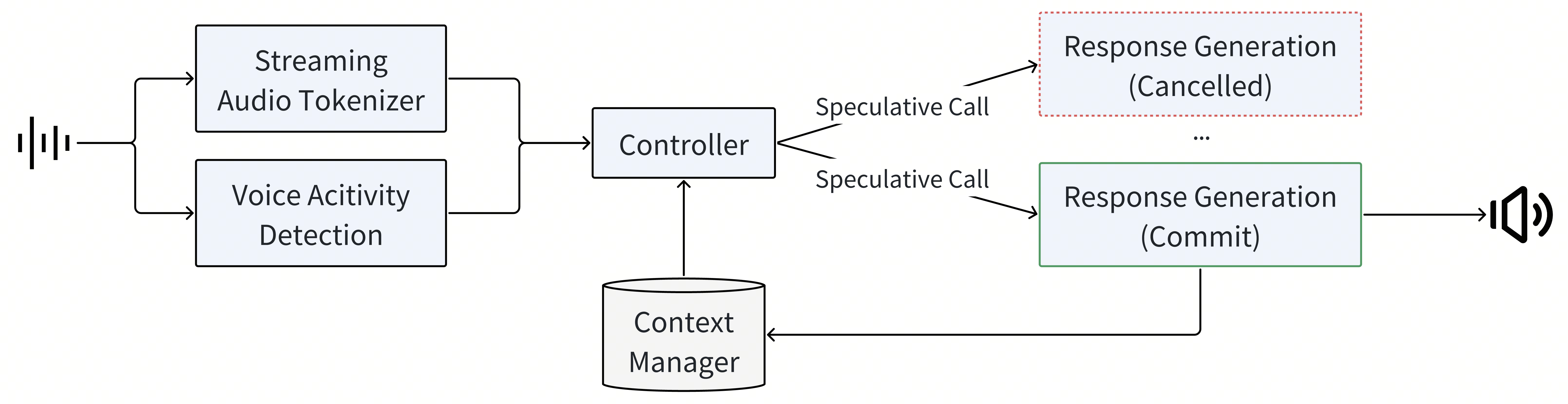}
  \caption{The architecture of the real - time inference pipeline aims to enable real-time interactions. When audio is input, it's processed concurrently by the streaming audio tokenizer and the voice activity detection module. The controller manages state transitions. A pause in user speech triggers speculative response generation, with multiple calls made but only one response committed. The context manager handles the conversation history in text format for continuity. Once the user finishes speaking, the system enters the reply state, commits a speculative response, and outputs audio. After that, it returns to the idle state for the next interaction.}
  \label{fig:inf_pipeline}
\end{figure}

\noindent To enable real-time interactions, we have designed an optimized inference pipeline as shown in Figure \ref{fig:inf_pipeline}. At its core, the Controller module manages state transitions, orchestrates speculative response generation, and ensures seamless coordination between critical subsystems. These subsystems include VAD for detecting user speech, the Streaming Audio Tokenizer for processing audio in real-time, the Step-Audio language model and Speech Decoder for processing and generating responses, and the Context Manager for preserving conversational continuity.

\paragraph{Speculative Response Generation}

To reduce interaction latency, the system preemptively generates speculative responses. This minimizes perceived delays and enhances responsiveness, though at the cost of occasional redundant computations when speculative responses are discarded. The system begins in the \texttt{Silence} state, awaiting user input. When the VAD detects active speech, the system transitions to the \texttt{UserSpeaking} state. During this state, the Streaming Audio Tokenizer begins converting audio into tokens. If the user momentarily pauses, the system enters the \texttt{UserPaused} state, where speculative response generation is triggered. By preemptively generating a response in anticipation of input completion, the system reduces latency when the conversation resumes. If the user resumes speaking, the speculative response is discarded. Once the system confidently determines that the user has finished speaking, it transitions to the \texttt{BotReplying} state, commits the most recent speculative response, and delivers its audio output. If interrupted by user speech, the system prioritizes the new input while maintaining conversational continuity. After completing its response, the system returns to the \texttt{Silence} state, ready for the next interaction. Empirical analysis shows that approximately 40\% of speculative responses are successfully committed. This mechanism reduces per-response latency by approximately 500ms compared to non-speculative methods.

\paragraph{Context Management}

Our system utilizes text transcription instead of raw audio tokens for historical context, as it provides a more compact representation (with an average text-to-audio token ratio of 1:14), improving performance, and enabling longer conversations with minimal impact on quality. ASR asynchronously transcribes user speech into text, maintaining an accurate and up-to-date conversation history.

\paragraph{Streaming Audio Tokenizer}

The input audio stream is processed through two parallel tokenizer pipelines, each employing fixed-duration segmentation. The resulting tokens are seamlessly merged into a single sequence with a 2:3 interleaving ratio. Without the streaming audio tokenizer, the inference time will be significantly slower, depending on the length of the audio input.


\section{Pretrain}
\label{chap:pretrain}
\subsection{Dataset}


\noindent Our multi-modal pretraining dataset integrates three major categories of data resources: \textbf{audio, text, and images}. The audio section comprises 1.1 trillion tokens of audio continuation data (approximately 7,300,000 hours), 113 billion tokens of TTS (Text-to-Speech) synthesized speech data (about 700,000 hours), 105 billion tokens of ASR (Automatic Speech Recognition) data (around 650,000 hours), and 350 billion tokens of audio-text alternating data (approximately 2,000,000 hours). The text data, amounting to 800 billion tokens, encompasses web documents, books, code, and proprietary materials. The image section includes 800 billion tokens of image-text paired/alternating data, sourced from web pages, books, and proprietary resources.

\subsection{Training Detail}
\noindent Step-Audio is a component of \textbf{Step-Omni}, which is designed to train a unified pretrained model for speech, image, and text. This training is based on a pretrained text model and image encoder for continued pretraining. The entire process is divided into three stages in total.
\begin{itemize}
\item \textbf{Stage1}: We expanded the vocabulary of the pretrained text model by adding 5,120 audio tokens and integrated a pretrained image encoder to form the \textbf{Step-Omni} model. During training, to ensure minimal loss of the text model's capabilities, the learning rate of the text model backbone is maintained at a low level (2e-5) throughout. However, the learning rates for the embedding and language model (LM) head are set five times higher than the backbone's to facilitate faster convergence of the newly added tokens. Meanwhile, the image encoder remains frozen during the entire training process. At this stage, audio, text, and image data are used in a 2:1:1 ratio, with audio data consisting solely of pure audio continuation tasks. 
\item \textbf{Stage2}: After training on 1.2T tokens in the stage1 phase, we incorporate audio-text interleaved data for further training, with a 1:1 ratio of audio continuation data to audio-text interleaved data. During this stage, the ratio of audio, text, and image data remains 2:1:1.
\item \textbf{Stage3}: After training on 800B tokens in the stage2 phase, we incorporate ASR and TTS data for further training. The ratio of audio continuation data, audio-text interleaved data, ASR data, and TTS data is set to 1:1:1:1. During this phase, the ratio of audio, text, and image data is adjusted to 4:3:3. Additionally, the learning rates for the embedding and LM head are synchronized with the backbone, utilizing a cosine schedule that decreases from 2e-5 to 5e-6.
\end{itemize}

\noindent We employ the same pre-training strategy across models of varying parameter scales.

\subsection{Training Infrastructure}

\noindent We train Step-Omni on thousands of H800 GPUs with 35\% Model Flops Utilization (MFU). Despite employing the standard optimizations such as tailored GPU kernels and communication overlap, we highlight two innovative approaches that further enhance our training efficiency.

\paragraph{Disaggregated Data Processing} The processing of multi-modality data in Step-Omni training is computationally intensive, often requiring substantial CPU resources to keep pace with the model training speed. Conventional implementations typically co-locate data processing tasks with training jobs, leading to significant interference between these tasks and ultimately slowing down the training process. To address this issue, we introduce StarWeaver, an RPC-based distributed data processing library. StarWeaver relocates CPU-intensive data pre-processing tasks to remote processes, thereby alleviating the computational burden on the GPU training side and enhancing overall training efficiency. StarWeaver also facilitates the enhancement of load balancing in the data-parallel dimension, as it serves as an ideal mechanism for redistributing data with global workload information.

\paragraph{Disaggregated Model Placement} For multi-modal models such as Step-Omni, training typically involves not only LLM but also modality encoders (e.g., vision encoder). Integrating these diverse components challenges the conventional assumption of training frameworks that the model is homogeneous and monolithic. This mismatch often results in suboptimal training efficiency. To address this issue, we propose disaggregated model placement that allocates dedicated resources and employs tailored parallelism strategies for each sub-model. This novel approach effectively minimizes pipeline bubbles caused by model heterogeneity, thereby achieving optimal training efficiency. Details can be found at~\citep{zhang2024disttrain}.

\subsection{Exploring Tokenizer for Audio Pretraining}
\noindent To achieve the unification of speech understanding and generation, we first explored the use of a speech tokenizer. Initially, we investigated the training approach using a single codebook. In our experiments, we found that when training the model using only semantic tokens, the next token prediction perplexity is relatively low, and the semantic coherence between the generated content and the preceding context is good. However, due to the significant loss of acoustic information from discarding too many semantic tokens, the subsequent audio restoration through the vocoder suffers severe degradation in terms of timbre and prosody, resulting in poor auditory quality. When only using linguistic tokens for training, the audio recovered by the vocoder from the model's continuation sounds good, but the next token prediction perplexity is very high, and the semantic coherence between the continuation and the preceding context is poor.
\begin{figure}[!ht]
    \centering
    \includegraphics[width=0.8\linewidth]{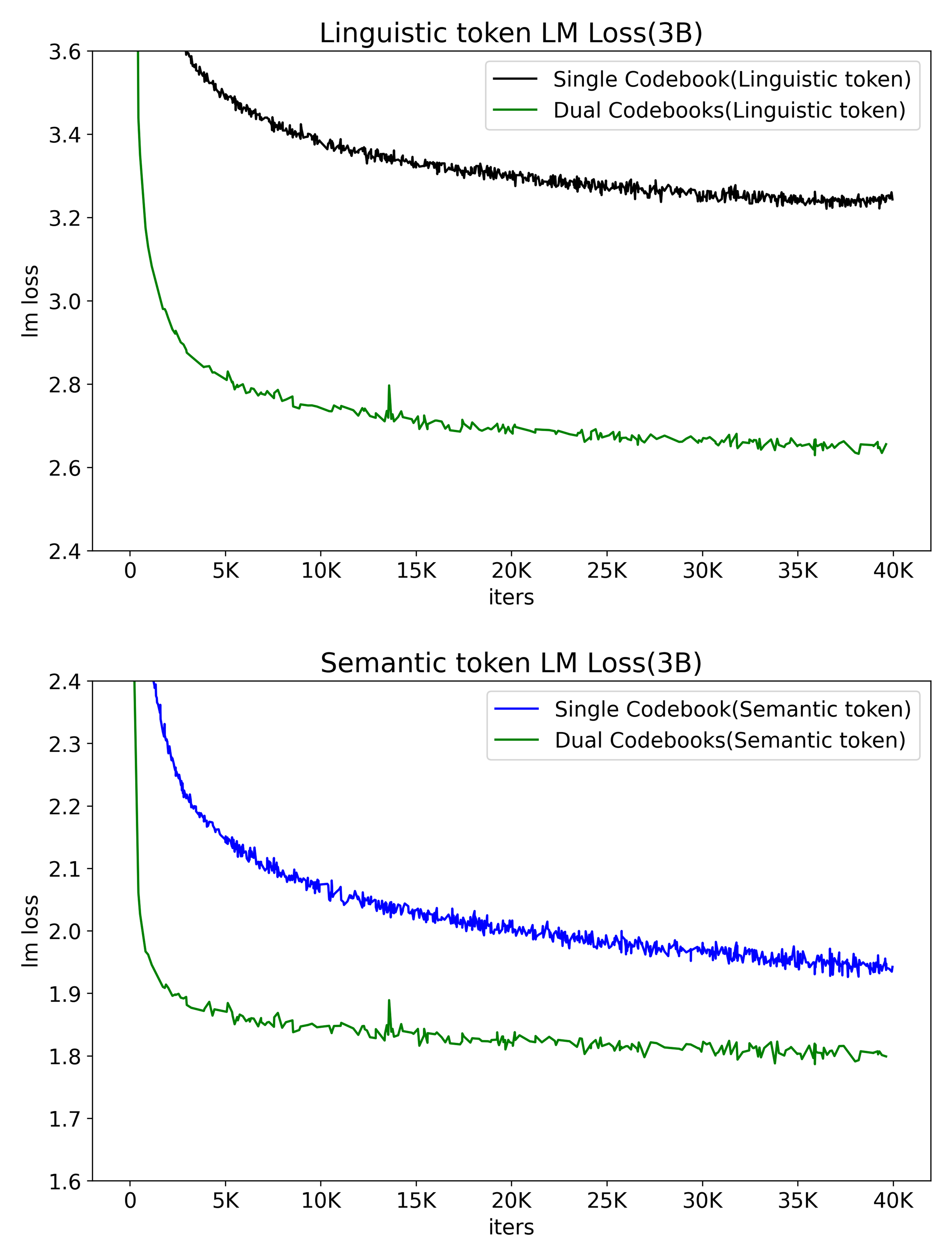}
    \caption{Training loss comparison between Dual-Codebook and Single Codebook Tokenizer.}
    \label{fig:dualvssingle}
\end{figure}
\noindent When training with interleaved semantic tokens and linguistic tokens, the semantic tokens ensure the semantic coherence of the continuation with the preceding context, while the linguistic tokens ensure the auditory quality of the reconstructed audio. Due to the mutual reference between semantic tokens and linguistic tokens, we observed that when using dual-codebook training, the next token prediction perplexity for both semantic tokens and linguistic tokens decreased compared to using a single codebook as shown in Figure \ref{fig:dualvssingle}. Notably, the decrease in next token prediction perplexity for semantic tokens was more significant. Furthermore, ASR ablation results indicated that the dual-codebook model achieved a lower character error rate (CER) on the ASR test set compared to the pure single-codebook model(see section \ref{sec:asr}). 

\bigskip
\noindent Furthermore, grouping and interleaving linguistic discrete tokens and semantic discrete tokens in a 2:3 ratio facilitates faster convergence of training loss. More importantly, extending the CosyVoice semantic tokens with linguistic tokens enhances the model's ability to understand and follow multi-turn history instructions and also mitigates issues such as unclear pronunciation and indistinct articulation, significantly boosting the performance of CosyVoice's single code.

\section{Post-Training}
\label{chap:posttrain}

\subsection{TTS}
\label{sec:tts}
\subsubsection{Dataset}
\label{sec:tts_sft_data}
\noindent High-quality speech data is crucial for TTS task, as it directly impacts the model's performance and the expressiveness of the generated speech. Language-specific data, dialect data, speaking styles, emotional data, and paralinguistic data are extremely scarce. Constructing such datasets demands substantial human and financial resources, and the process generally spans an extended period. 

\bigskip
\noindent To address this gap, we present the first novel synthetic data-driven framework for TTS systems, comprising three key components:
\begin{itemize}
\item First, we employ a Step-2~\citep{step2} LLM to generate linguistically diverse and semantically rich textual content.
\item Second, we selected a pre-trained Step-Audio model checkpoint incorporating audio-token cooldown mechanisms, which enables direct generation of speaker-specific, language-dependent, and dialect-aware audio data.
\item Third, we developed an Audio-Edit Model by fine-tuning the aforementioned checkpoint, specifically designed to generate nuanced emotional expressions and diverse speaking styles. This model architecture allows for precise control over paralinguistic features while maintaining speaker consistency.
\end{itemize}
\begin{figure}[htbp]
  \centering
  \includegraphics[width=1.0\textwidth]{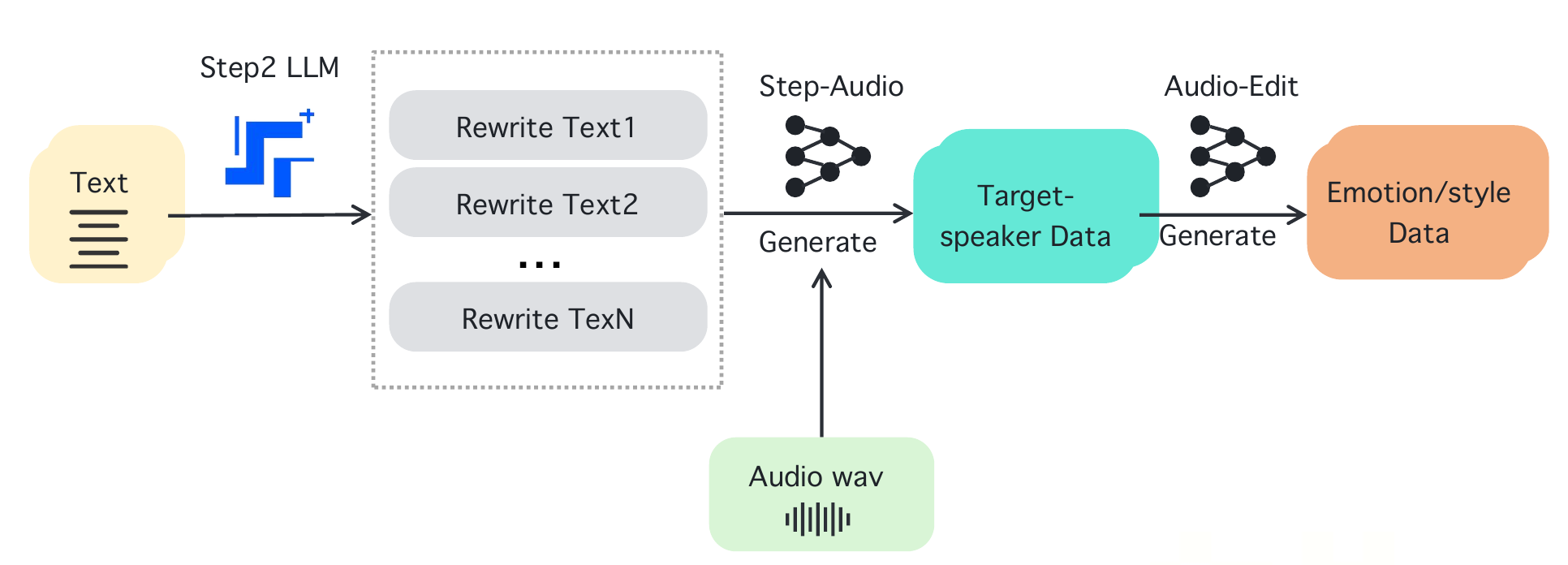}
  \caption{The process starts with text input which is processed by a Step-2 LLM to generate multiple rewritten texts. Then, a Step-Audio model generates target-speaker data using the rewritten texts and existing audio wav data. Finally, an Audio-Edit model refines the data to produce emotion/style data, addressing the scarcity of high - quality speech data in TTS tasks.}
  \label{fig:pipeline}
\end{figure}

\paragraph{Language and Dialect} Leveraging the robust continuation ability of Step-Audio, which has been trained on large volumes of speaker and language data, we generate target-speaker, language and dialect data. The text-based LLM Step-2 is used to translate and rewrite chat text to conform to the grammar and style of the target language or dialect. We collect audio recordings and texts from native speakers as prompt audio and text, and then, using the format 
[system prompt; prompt text; target text; prompt code; target code]
along with the corresponding text, we use Step-Audio for audio continuation generation. This method allows for the quick creation of a large amount of native-speaker data for the target language and dialect with only a small quantity of high-quality seed data.

\paragraph{Emotion and Speaking Styles} Emotion and speaking style data have been challenging to deal with because of the difficulty in both differentiating and defining emotion categories and their respective intensities, as well as the complexity associated with accurately describing and recording various style types. To address this, an Audio-Edit model-based approach is proposed. It ingeniously converts complex emotion and style descriptions into a comparative pair data construction format. Step-2 is used to rewrite chat text with specific emotions and styles. Normal and emotional speech samples from the same speaker with identical text are collected, and Step-Audio is used for cloning and continuation generation to create (text, neutral audio token, emotion and style audio token) data. Only the (neutral audio token, emotion and style audio token) pairs are used to perform SFT on the audio cooldown pretrain model to get the Audio-Edit model. Using this model, neutral-style speech can be input to generate emotion or style enhanced audio, and data with different emotion or style intensities can be produced iteratively. 

\paragraph{Singing and RAP}
We construct a paired dataset of lyrics and vocal segments through three stages: (1) Collecting 10,000+ hours of singing / RAP tracks with LyRiCs-format timestamps; (2) Extracting dry vocals using Demucs~\citep{rouard2022hybrid} and removing silent regions via Voice Activity Detection(VAD); (3) Segmenting audio using LyRiCs timestamps and aligning lyrics with audio segments. For data cleaning, We performed three steps: (1) RAP Separation: we isolated pure RAP segments by retaining those with higher speech rates and using a genre classification model to identify hip-hop clips; (2) Audio Quality Filtering: Utilizing noise detection and speaker diarization, we preserved low-noise, single-speaker segments; (3) Alignment Verification: To address misalignment due to inaccurate LyRiCs timestamps, we computed the Character Error Rate (CER) between transcribed speech and ground-truth lyrics, discarding misaligned segments. Ultimately, the total length of the retained audio segments constituted 17.8\% of the original song durations. This dataset supports dual training objectives: the LLM learns to map lyrics to linguistic and semantic tokens, while the speech decoder decodes these tokens into high-fidelity vocals in precise tunes. 

\paragraph{Target Speaker}
Supporting multiple languages or dialects for a target speaker is challenging through model generalization of foundational language and dialect data, as it often fails to achieve the level of a native speaker. To mitigate this issue, we employ dual codes extracted from audio generated by native speakers with timbre and prosody similar to the target speaker. These dual codes are combined with the target speaker's prompt audio to regenerate new audio, from which dual codes are then extracted again. Through this straightforward procedure, the target speaker's speech in new languages and dialects becomes more akin to that of a native speaker.

\bigskip
\noindent 
Quality assessment of data constitutes a critical component in our synthetic data framework. To ensure the reliability and validity of both seed and synthesized data, we have implemented a comprehensive evaluation system incorporating multiple objective metrics: ASR accuracy, Voice Activity Detection (VAD) performance, Speaker Diarization precision, Emotion recognition consistency, and Deep Noise Suppression (DNS) effectiveness. This multi-dimensional quality control mechanism guarantees the robustness and practical utility of generated synthetic data.

\subsubsection{Training Detail}
\noindent In contrast to conventional TTS systems that emphasize fine-grained control over speaker characteristics, emotional expression, linguistic features, and stylistic elements, our approach adopts the chat-based paradigm and training methodology of LLMs. This strategic alignment significantly enhances system flexibility while simultaneously establishing a scalable framework to support future model and data expansion, thereby addressing the critical challenges of scalability in speech synthesis systems.

\paragraph{Supervised Fine-Tuning Format} The sft format comprises three essential components: the system prompt, the human input, and the assistant response, structured in a two-turn dialogue configuration. Within this format, the system prompt serves as the foundational element for specifying the speaker attributes and defining the supported instruction tags. The human input and the assistant response components are specifically designed to handle the textual content and the dual-codebook representations respectively. The text and audio tokens from the first round can be utilized to maintain the in-domain speaker's timbre and style consistency, as well as to enable out-domain zero-shot cloning. 

\paragraph{Instruction Tags}  Instruction tags are classified into two distinct categories: descriptive tags and comparative tags. Descriptive tags are utilized for controlling aspects such as language, dialect, vocal, and style, while comparative tags are employed for hierarchical distinctions in emotion and speed control. The data for descriptive tags are generated using the Step-Audio model clone, supporting languages and styles including Japanese, Korean, Cantonese, Sichuan dialect, cute voice, RAP, and singing. The data for comparative tags are generated using the Audio Edit model, supporting emotions such as happiness, anger, sadness, and speed variations like fast and slow, each divided into five hierarchical levels.

\bigskip
\noindent We employ the SFT data as outlined in Section \ref{sec:tts_sft_data}. And utilize a 3-billion parameter model, training it for one epoch with an initial learning rate of $2 \times 10^{-5}$ .The learning rate is adjusted using a cosine decay strategy, with a lower bound set at $2 \times 10^{-6}$. 

\subsection{AQTA}

\begin{figure}[htbp]
  \centering
  \includegraphics[width=1.0\textwidth]{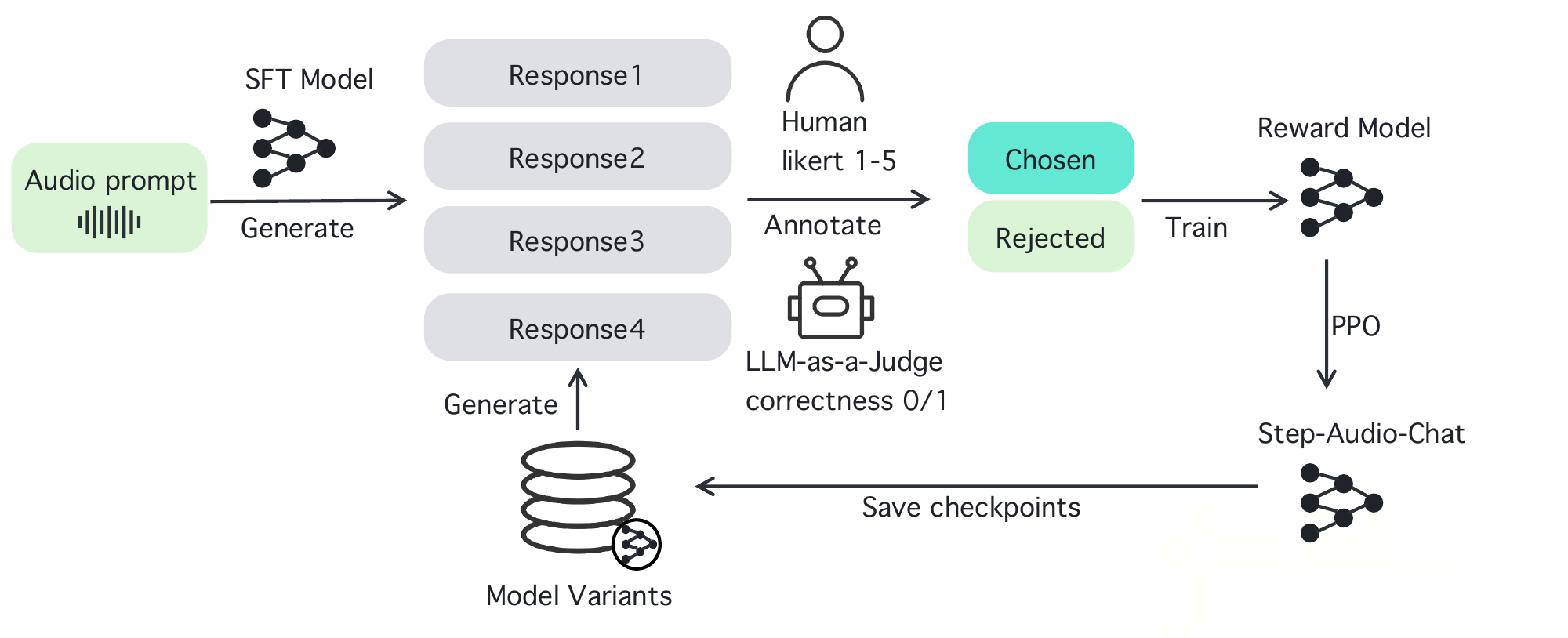}
  \caption{At each training iteration, we collect multiple responses from different versions of the model. Then, through manual scoring, as well as evaluated by a LLM, high-quality pairs are selected to train the reward model. Finally, we use PPO algorithm to train the final Step-Audio-Chat model.}
  \label{fig:rlhf}
\end{figure}

We applied Reinforcement Learning from Human Feedback (RLHF) for the AQTA task, leading to the creation of the Step-Audio-Chat model, as depicted in Figure \ref{fig:rlhf}.

\subsubsection{SFT dataset}
\label{sec:sft_data}
\paragraph{Data Types}

We categorized the SFT data into several types based on the nature of the input (Q) and output (A):

\begin{itemize}
    \item \textbf{TQTA}: This type includes a substantial amount of text-based Question-Answer (QA) data. 
    \item \textbf{AQTA}: This type consists of audio inputs paired with textual outputs. 
    \item \textbf{TAQTA}: This type is designed to enhance the consistency between text and speech. Here, the text Q serves both as an input (not contributing to the loss calculation) and as an output (contributing to the loss calculation). 
    \item \textbf{Other Types}: These include audioQ-audioA(AQAA), visionQ-audioQ-textA(VAQTA) and so on. These types are included to provide additional diversity and complexity to the training data, further improving the model's robustness.
\end{itemize}
To enhance speech recognition capabilities, we have incorporated additional training data annotated in ASR format alongside existing datasets. These ASR-formatted resources contain detailed transcriptions of speech signals, enabling models to better interpret phonetic patterns and linguistic nuances. The integration of such supplementary ASR-annotated data strengthens model robustness against acoustic variations including regional accents, speaking rate fluctuations, and ambient noise interference. 

\paragraph{Data Processing}

To optimize the SFT data for effective model training, we implemented the following processing steps:

\begin{itemize}
    \item \textbf{Single-Turn Data Modification:} For single-turn interactions, we applied text length filtering to the inputs. This is because real-world user speech inputs are often concise. Additionally, we modified the outputs to adopt a more conversational text style, enhancing the speech model's human-like qualities and avoiding rigid, verbose, or overly structured responses.
    \item \textbf{Multi-Turn Data Processing:} For multi-turn interactions, we replaced the speech inputs from previous turns with their corresponding text transcriptions. Only the speech input from the final turn was retained. Furthermore, only the response from the final turn was considered in the loss calculation, focusing the model's training on generating accurate and relevant responses to the most recent input.
\end{itemize}
Through this systematic approach to SFT data construction and processing, we aimed to create a diverse and representative dataset that would enable our speech model to achieve superior performance in real-world scenarios, delivering natural, coherent, and contextually appropriate responses to user inputs.

\subsubsection{Supervised Fine-Tuning Details}
\noindent We use SFT data as described in section \ref{sec:sft_data}. And the model is finetuned for 1 epoch with learning rate from $5.656 \times 10^{-5}$ to $5.656 \times 10^{-6}$.

\subsubsection{Reward Model dataset}
\label{sec:rm_data}

\paragraph{TQTA Preference Data Construction}
We collected human preference data generated by the TQTA model~(e.g., Step-1 and Step-2) and removed categories that were less distributed in speech dialogues, such as code and mathematics. We mainly retained categories such as daily conversations, role-playing, safety and instruction following.
\paragraph{AQTA Preference Data Construction}
For the fine-tuning dataset, we first collected real audio prompts from users and sampled four responses using the SFT model. We then constructed chosen/rejected pairs by having human annotators rate these four responses on a scale of 1 to 5, based on the criteria of instruction following, conversational naturalness and safety. In addition to these artificially generated labels, we also employed the LLM-as-a-Judge method to score the model's responses to objective questions and create corresponding chosen/rejected pairs based on responses' correctness.
To mitigate the pattern bias associated with ``deaf hacking'' as described in \ref{sec:ppo}, we employed the hacked PPO model to generate responses for input audio with clear audio prompts. If the responses exhibited hacking behavior, we constructed them as rejected responses. This process aimed to eliminate the pattern bias caused by the exclusive presence of ``deaf hacking'' as chosen responses in the training data of the reward model.

\subsubsection{Reward Model Training Details}
\label{rm_training}
\noindent We implement a two-stage approach for reward model training: TQTA single-modal preference model pretraining, followed by AQTA cross-modal fine-tuning. The model is fine-tuned for 1 epoch on TQTA and 1 epoch on AQTA. The learning rate is adjusted using a cosine decay strategy, initialized at $1.24 \times 10^{-5}$ with a lower bound set at $6 \times 10^{-6}$.
\bigskip

\noindent The reward model training initializes from a SFT model and proceeds through the two-stage training using the Bradley-Terry loss~\citep{Bradley1952RankAO}, achieved a pair-wise accuracy of 70.51\% on the human preference test set. 

\subsubsection{PPO dataset}
\noindent For the PPO training data, we used the same prompt seeds as those employed in the AQTA fine-tuning stage of the reward model.

\subsubsection{PPO Training Details}
\label{sec:ppo}
\noindent After obtaining the reward model, we employ the PPO~\citep{schulman2017proximalpolicyoptimizationalgorithms} algorithm to train speech large language model. During the RLHF training stage, the critic model is warmed up with an initial 80 training steps ahead. We employ a PPO clip threshold of $\epsilon = 0.2$ and an initial learning rate of $1 \times 10^{-6}$, which decays using a cosine strategy, with a minimum learning rate of $2 \times 10^{-7}$. Additionally, we set the KL penalty coefficient to $\beta = 0.05$.
\bigskip

\noindent Unlike the reward hacking observed in the RLHF training of TQTA models, we found that a reward model trained exclusively on human-annotated AQTA preference data exhibited a ``deaf hacking'' phenomenon (i.e., the reward model assigned high reward to responses containing phrases like ``I didn't hear clearly'' regardless of input audio clarity, unintentionally reinforcing deaf hacking patterns during RLHF training). We attributed this issue to a pattern bias in the reward model's training data, which exclusively featured ``deaf hacking'' pairs: the model responded with ``I didn’t hear clearly'' to unclear or semantically incomplete prompts as chosen responses but lacked such responses to clear and semantically complete prompts as rejected ones. To mitigate this bias, we constructed corresponding data as mentioned in Section \ref{sec:rm_data}. We also plan to introduce rule-based rewards during RLHF training to eliminate ``deaf hacking'' in future work.


\section{Evaluation}
\subsection{Benchmark Design}
\label{sec:bmk_design}
\noindent We have created a new benchmark, named StepEval-Audio-360\footnote{\url{https://huggingface.co/datasets/stepfun-ai/StepEval-Audio-360}}, following a series of rules. In terms of design principles, this benchmark aims to fill the gaps in the evaluation of multi-modal speech interaction, systematically identify the strengths and weaknesses of the the models, and attach importance to user experience and safety. For data collection, real user recordings are used in combination with public corpora. Meanwhile, strict control is exercised on audio quality and semantic annotation to ensure compliance with privacy. 
\bigskip

\noindent The evaluation dimensions mainly cover language proficiency, emotional intelligence, logical reasoning, creativity, multi-instruction following, role-playing, safety, etc. Demographic differences (age/ gender/ dialect), environmental conditions (noise level/ microphone type), and prosodic features (speech rate/ pronunciation pattern) are also taken into account. The indicator system architecture combines quantitative analysis, uses scripts for automatic verification of indicators such as accuracy rate and repetition rate, and also involves evaluation by large language models and human. In addition, quarterly updates of the benchmark are carried out to avoid falling behind, and adjustments are made in light of user feedback.

\subsection{Results}

\subsubsection{ASR}
\label{sec:asr}

\noindent We conducted validation experiments with a 3B model to compare the performance of Semantic Code and Dual-Code on ASR (Automatic Speech Recognition) tasks. While keeping the same amount of audio training data, the Character Error Rate (CER) of the Dual-Code approach improved from 25.5 to 18.4. This demonstrates that the Dual-Code method significantly enhances performance on ASR tasks. 
\bigskip


\noindent We tested the performance of the model at two stages: first, the pre-trained model (Step-Audio Pretrain) ; and second, the chat model (Step-Audio-Chat) , after the alignment of human preference, using the system prompt ``\begin{CJK*}{UTF8}{gbsn} 请记录下你所听到的语音内容。 \end{CJK*}". The evaluation datasets include Aishell1, Aishell2 ios, Wenetspeech test-net, Wenetspeech test-meeting, Librispeech test-clean, and Librispeech test-other. For the evaluation metrics, we employ the character error rate (CER) for Chinese and word error rate (WER) for English. We systematically compared the following two categories of mainstream large speech models:
\begin{itemize}
\item Hidden feature modeling: Whisper Large-v3~\citep{radford2023robust}, Qwen2-Audio~\citep{chu2024qwen2audiotechnicalreport}, MinMo~\citep{chen2025minmomultimodallargelanguage}, LUCY~\citep{gao2025lucylinguisticunderstandingcontrol}; 
\item Discrete audio token modeling: Moshi~\citep{2024moshispeechtextfoundationmodel}, GLM-4-voice~\citep{zeng2024glm4voiceintelligenthumanlikeendtoend}, Step-Audio.
\end{itemize}
\begin{table}[htbp]
    \centering
    \caption{ASR result comparison}
    \label{tab:comparison}
    \footnotesize
    \resizebox{\textwidth}{!}{
    \begin{tabular}{lcccccccccc}
        \toprule
        \small
        & \multicolumn{4}{c}{Hidden Feature Modeling} & \multicolumn{5}{c}{Discrete Audio Token Modeling} \\
        \cmidrule(lr){2-5} \cmidrule(lr){6-10}
        & \makecell[c]{Whisper\\ Large-v3} & Qwen2-Audio & MinMo  & LUCY & Moshi & \makecell[c]{GLM-4-voice\\ Base} & \makecell[c]{GLM-4-voice\\ Chat}  & \makecell[c]{Step-Audio \\Pretrain} & \makecell[c]{Step-Audio\\Chat} \\
        \midrule
        Aishell-1 & 5.14 & 1.53 & - & 2.4 & - & 2.46 & 226.47  & \textbf{0.87} & 1.95  \\
        Aishell-2 ios & 4.76 & 3.06 & \textbf{2.69}  & - & - & - & 211.3  & 2.91 & 3.57\\
        Wenetspeech test-net & 9.68 & 7.72 & \textbf{6.64}  & 8.78 & - & - & 146.05  & 7.62 & 8.75 \\
        Wenet test-meeting & 18.54 & 8.4 & \textbf{7.6}  & 10.42 & -  & - & 140.82 & 7.78 & 9.52\\
        Librispeech test-clean & 1.9 & \textbf{1.6} & \textbf{1.6} & 3.36 & 5.7   & 2.82 & 75.39 & 2.36 & 3.11 \\
        Librispeech test-other & 3.65 & \textbf{3.6} & 3.82  & 8.05 & -  & 7.66 & 80.3  & 6.32 & 8.44 \\
        \midrule
        AVG & 7.28 & \textbf{4.32} & -  & - & - & - & 146.74  & 4.64 & 5.89 \\
        \bottomrule
    \end{tabular}
    }
\end{table}

\noindent The specific results are shown in Table \ref{tab:comparison}. Among the audio token-based speech models, Step-Audio Pretrain achieved the best performance with an average CER of 4.64. Compared to the hidden feature models, Step-Audio Pretrain outperformed Whisper Large-v3 and achieved comparable results to Qwen2-Audio and MinMo, particularly on the clean test sets of Aishell1, Aishell2 and Librispeech test-clean, where Step-Audio Pretrain (average CER 2.05 ) was very close to Qwen2-Audio (average CER 2.06). This indicates that Step-Audio, through its dual-codebook compression strategy, has effectively preserved semantic information during the discretization of speech representation. 

\bigskip
\noindent Additionally, we compared the ASR capabilities of the final conversational models. When testing the GLM-4-voice Chat model, we experimented with various prompts and ultimately selected the prompt ``\begin{CJK*}{UTF8}{gbsn} 请写下你听到的语音内容： \end{CJK*}" for evaluation. However, the model still struggled to follow instructions effectively. In contrast, the Step-Audio Chat model achieved an average CER of 5.89, maintaining strong performance, which reflects its robust ability to follow instructions.

\begin{table}[hbt]
    \centering
    \caption{Performance comparison of content consistency across Step-Audio, GLM-4-Voice, and MinMo.}
    \label{tab:cer_wer}
    \resizebox{0.6\textwidth}{!}{
    \begin{tabular}{lcccccc}
        \toprule
        \multirow{2}{*}{Model} & \multicolumn{1}{c}{test-zh} & \multicolumn{1}{c}{test-en} \\
        \cmidrule(lr){2-3} \cmidrule(lr){4-5} 
        & CER (\%) $\downarrow$  & WER (\%) $\downarrow$ \\
        \midrule
        GLM-4-Voice\textsuperscript & 2.19  & 2.91   \\
        MinMo\textsuperscript &  2.48  & 2.90  \\
        \midrule
        \textbf{Step-Audio}\textsuperscript & \textbf{1.53} & \textbf{2.71} \\
        \bottomrule
    \end{tabular}
    }
\end{table}
\begin{table}[ht]
    \centering
    \caption{Results of Step-Audio-TTS-3B and recent LLM-Based TTS models on the SEED test sets. Step-Audio-TTS-3B-Single denotes dual-codebook backbone with single-codebook vocoder. Step-Audio-TTS denotes 130 Billion parameter version of Step-Audio-TTS.}
    \vspace{10pt} 
    \label{tab:cer_ss_compare}
    \resizebox{\textwidth}{!}{
    \begin{tabular}{lcccccc}
        \toprule
        \multirow{2}{*}{Model} & \multicolumn{2}{c}{test-zh} & \multicolumn{2}{c}{test-en} \\
        \cmidrule(lr){2-3} \cmidrule(lr){4-5} 
        & CER (\%) $\downarrow$ & SS $\uparrow$ & WER (\%) $\downarrow$ & SS $\uparrow$ \\
        \midrule
        FireRedTTS & 1.51 & 0.630 & 3.82 & 0.460 \\
    MaskGCT & 2.27 & 0.774 & 2.62 & 0.774 \\
    CosyVoice & 3.63 & 0.775  & 4.29 & 0.699 \\
    CosyVoice 2 & 1.45 & 0.806 & 2.57 & 0.736 \\
    CosyVoice 2-S & 1.45 & 0.812  & 2.38 & 0.743 \\ \midrule
    \textbf{Step-Audio-TTS-3B-Single} & 1.37 & 0.802 & 2.52 & 0.704 \\
    \textbf{Step-Audio-TTS-3B} & \textbf{1.31} & 0.733 & \textbf{2.31} & 0.660 \\ 
    \textbf{Step-Audio-TTS} & \textbf{1.17} & 0.73 & \textbf{2.0} & 0.660 \\ 
        \bottomrule
    \end{tabular}
    }
\end{table}
\begin{table}[hbt]
\centering
\caption{Performance comparison of Dual-codebook Resynthesis with Cosyvoice.}
\label{tab:dualvssingle}
    \vspace{10pt} 
    \resizebox{0.75\textwidth}{!}{
\begin{tabular}{lcccccc}
\toprule
\multirow{2}{*}{Token} & \multicolumn{2}{c}{test-zh} & \multicolumn{2}{c}{test-en} \\
\cmidrule(lr){2-3} \cmidrule(lr){4-5} 
& CER (\%) $\downarrow$ & SS $\uparrow$ & WER (\%) $\downarrow$ & SS $\uparrow$ \\
\midrule
Groundtruth & 0.972 & - & 2.156 & -  \\
CosyVoice & 2.857 & \textbf{0.849} & 4.519 & \textbf{0.807}  \\
Step-Audio-TTS-3B & \textbf{2.192} & 0.784 & \textbf{3.585} & 0.742  \\

\bottomrule
\end{tabular}
}
\end{table}

\subsubsection{TTS}
\noindent To evaluate the TTS performance of Step-Audio, we utilized the SEED TTS test dataset \citep{anastassiou2024seedttsafamilyofhighquality}. The model was prompted to reproduce the original input text. Although this approach cannot guarantee the consistency between input and output contents, we believe that erroneous outputs also objectively reflect the model's ability to follow instructions. Therefore, we employed Paraformer~\citep{gao2023paraformerfastaccurateparallel} and Whisper-Large-V3~\citep{radford2023robust}  to calculate the error rate for Chinese and English, respectively. The results are summarized in Table \ref{tab:cer_wer}. Step-Audio achieved the best CER and WER performance among open-source spoken models. 

\bigskip
\noindent For the TTS task, we evaluated the performance of open-source TTS models on the SEED test sets using Paraformer~\citep{gao2023paraformerfastaccurateparallel} and Whisper-Large-v3~\citep{radford2023robust}. And the ERes2Net~\citep{anenhancedres2netwithlocal} model is employed to evaluate speaker similarity. The results are summarized in Table \ref{tab:cer_ss_compare}.  Our 3-Billion parameter version of Step-Audio-TTS-3B with dual-codebooks achieved SoTA results in terms of CER and WER among the open-source models, while also demonstrating highly competitive similarity scores. Notably, scaling the LLM to 130 billion parameters yielded substantial improvements in both CER and WER, suggesting the potential benefits of further scaling both synthetic data and model parameters.

\bigskip
\noindent To evaluate the benefits of the dual-codebook approach, we compare the resynthesis quality of various tokens including: single cosyvoice codebook~\citep{du2024cosyvoicescalablemultilingualzeroshot}, and dual-codebook, on the aforementioned SEED test sets. The test speech is first quantized into discrete tokens and then reconstructed into waveforms using the token2wav process. We assess the quality of the tokens in terms of speech intelligibility and speaker similarity. The results of this comparison are presented in Table \ref{tab:dualvssingle} which shows that the dual-codebook approach achieves a lower CER while maintaining an acceptable level of degradation of speaker similarity.

\subsubsection{AQTA Chat}
\noindent As illustrated in the Table \ref{innerbmk-VoiceChatPerformance}, we compared the real-time dialogue performance of Step-Audio and open-source models use StepEval-Audio-360 benckmark mentioned in section \ref{sec:bmk_design}. The scores of each metric are automatically assessed by GPT-4o. And the average scores are shown in the table while the best results are shown in bold. Since the main content of this benchmark is in Chinese, and Moshi has almost no understanding of Chinese, the results from Moshi are marked with ``*'' and should be considered for reference only. The results indicated that Step-Audio-Chat demonstrated superior performance in real-time dialogue. 
\begin{itemize}
	\item \textbf{Factuality}: Step-Audio-Chat achieves the highest score of 66.4\%, significantly surpassing the other models. This indicates that Step-Audio-Chat provides responses that are more accurate and reliable, adhering closely to the factual information.
	\item \textbf{Relevance}: With a score of 75.2\%, Step-Audio-Chat also excels in relevance, demonstrating its ability to generate contextually appropriate and meaningful responses to user queries.
	\item \textbf{Chat Score:}  With the highest overall chat score (4.11), Step-Audio-Chat provides a superior overall voice chat experience.  This score, ranging from 1 (lowest) to 5 (highest), represents a comprehensive evaluation by GPT-4o based on the textual input and output of the conversations.
\end{itemize}
\begin{table}[ht]
	\centering
	\caption{Comparison of fundamental capabilities of voice chat on the StepEval-Audio-360.}
	\label{innerbmk-VoiceChatPerformance}
	\resizebox{0.8 \textwidth}{!}{
		\begin{tabular}{lcccc}
			\toprule
			Model & Factuality (\%) $\uparrow$ & Relevance (\%) $\uparrow$ & Chat Score$\uparrow$\\
			\midrule
			GLM4-Voice & 54.7 & 66.4 &  3.49\\
			Qwen2-Audio & 22.6 & 26.3 &  2.27\\
			Moshi\textsuperscript{*} & 1.0 & 0 & 1.49\\
			\midrule
			\textbf{Step-Audio-Chat}& \textbf{66.4} & \textbf{75.2} & \textbf{4.11} \\
			\bottomrule
		\end{tabular}
	}
	
\end{table}
\begin{table}[htbp]
	\centering
	\caption{Performance comparison of voice chat models on public benchmarks. Llama Question, Web Questions, and TriviaQA are publicly available datasets, while the audio versions of ComplexBench and HSK-6 are newly constructed from public text corpora in this study from publicly available text corpora. In the absence of an official test set, TriviaQA's results are for reference only and marked with ``*''. Qwen2-Audio and Step-Audio-chat are obtained via local inference while others (Moshi, Freeze-Omni, LUCY, MinMo) are sourced from their original publications. Notably, GLM4-Voice's scores on HSK-6 and ComplexBench are also generated through local inference.}
	\label{openbmk-VoiceChatPerformance}
	\resizebox{\textwidth}{!}{
		\begin{tabular}{lccc|cc}
			\toprule
			Model & Llama Question & Web Questions & TriviaQA\textsuperscript{*} & ComplexBench& HSK-6\\
			\midrule
			GLM4-Voice & 64.7 &32.2  & 39.1 & 66.0& 74.0\\
			Moshi & 62.3 & 26.6 & 22.8 & - & - \\
			Freeze-Omni & 72.0 & 44.7 & 53.9 &- &- \\
			LUCY& 59.7 & 29.3 & 27.0 & -&- \\
			MinMo & 78.9 & 55.0 & 48.3 &- &- \\
			\midrule
			Qwen2-Audio & 52.0 & 27.0 & 37.3 &54.0 & -\\
			\textbf{Step-Audio-Chat}&  \textit{\textbf{81.0}} & \textbf{75.1} & \textbf{58.0} & \textbf{74.0}& \textbf{86.0} \\
			\bottomrule
		\end{tabular}
	}
\end{table}


\bigskip
\noindent To further contextualize Step-Audio-Chat's performance, we conducted evaluations on several publicly available datasets. Among them, Web Questions, Llama Questions, and TriviaQA are knowledge-based question-answering datasets, while ComplexBench and the listening comprehension section of the Hanyu Shuiping Kaoshi (HSK-6) are comprehensive tests.  We utilized the full test sets for Web Questions and Llama Questions. For TriviaQA, which lacks ground truth answers for the official test set, we constructed a 1000-sample test set from the development set.  This subset comprised 500 samples each from the Wikipedia and Web verified sets, supplemented with data from the Web and Wikipedia dev sets.  Due to this non-standard test set, TriviaQA results should be considered preliminary and for reference purposes only.  While Llama Questions provided spoken questions, we used TTS to generate spoken versions of the questions for Web Questions, TriviaQA, ComplexBench and HSK-6, enabling an Audio Question-Text Answer (AQTA) evaluation.
\bigskip

\noindent Performance data for Moshi~\citep{2024moshispeechtextfoundationmodel}, Freeze-Omni~\citep{wang2024freezeomnismartlowlatency}, LUCY~\citep{gao2025lucylinguisticunderstandingcontrol}, and MinMo~\citep{chen2025minmomultimodallargelanguage} were taken from their respective publications.  Step-Audio-Chat and Qwen2-audio~\citep{chu2024qwen2audiotechnicalreport} results were obtained through local API inference.  For LUCY, we report the best results from its stage 2 (S2) and stage 3 (S3) training phases, as presented in the original publication.  GPT-4o was used to assess the accuracy of the model's textual responses against the original textual questions. Table \ref{openbmk-VoiceChatPerformance} presents the average accuracy scores, demonstrating that Step-Audio-Chat achieved the highest accuracy across all open-source benchmarks. The accuracy score for Step-Audio-Chat on the Llama Question dataset is italicized to indicate that we corrected several errors in the evaluation set. In addition to the automated metrics, we conducted a human evaluation as shown in Figure \ref{fig:huamn_eval}. The results corroborate the automated evaluation, showing a clear and consistent advantage for Step-Audio across all assessed dimensions.

\subsection{Instruction Following}
\noindent Audio instruction following reflects the model's capability to generate accurate audio and textual content in response to input instructions. Consequently, we have developed a benchmark for Audio Instruction Following, encompassing categories such as Languages, Role-playing, Singing / RAP, and Voice Control. The evaluation comprises the accuracy of instruction following and quality of generated speech, both assessed using a 1-5 Mean Opinion Score (MOS) scale. As demonstrated in Table \ref{table:instructionFollow}, Step-Audio-Chat shows competitive results in audio instruction following and audio quality respectively.
\begin{table}[htbp]
	\centering
	\caption{Performance comparison of audio instruction following between GLM4-Voice and Step-Audio-Chat}
	\label{table:instructionFollow}
	\resizebox{\textwidth}{!}{
		\begin{tabular}{lcccc}
			\toprule
                \multirow{2}{*}{Category} & \multicolumn{2}{c}{Instruction Following} & \multicolumn{2}{c}{Audio Quality} \\
                \cmidrule(lr){2-3} \cmidrule(lr){4-5} 
			&GLM-4-Voice&Step-Audio&GLM-4-Voice&Step-Audio\\
                \midrule
                Languages&1.9&3.8&2.9&3.3\\
                Role-playing&3.8&4.2&3.2&3.6\\
                Singing / RAP&2.1&2.4&2.4&4\\
                Voice Control&3.6&4.4&3.3&4.1\\
			\bottomrule
		\end{tabular}
	}
\end{table}
\subsection{Toolcall}
\begin{figure}[htbp]
  \centering
  \includegraphics[width=1\textwidth]{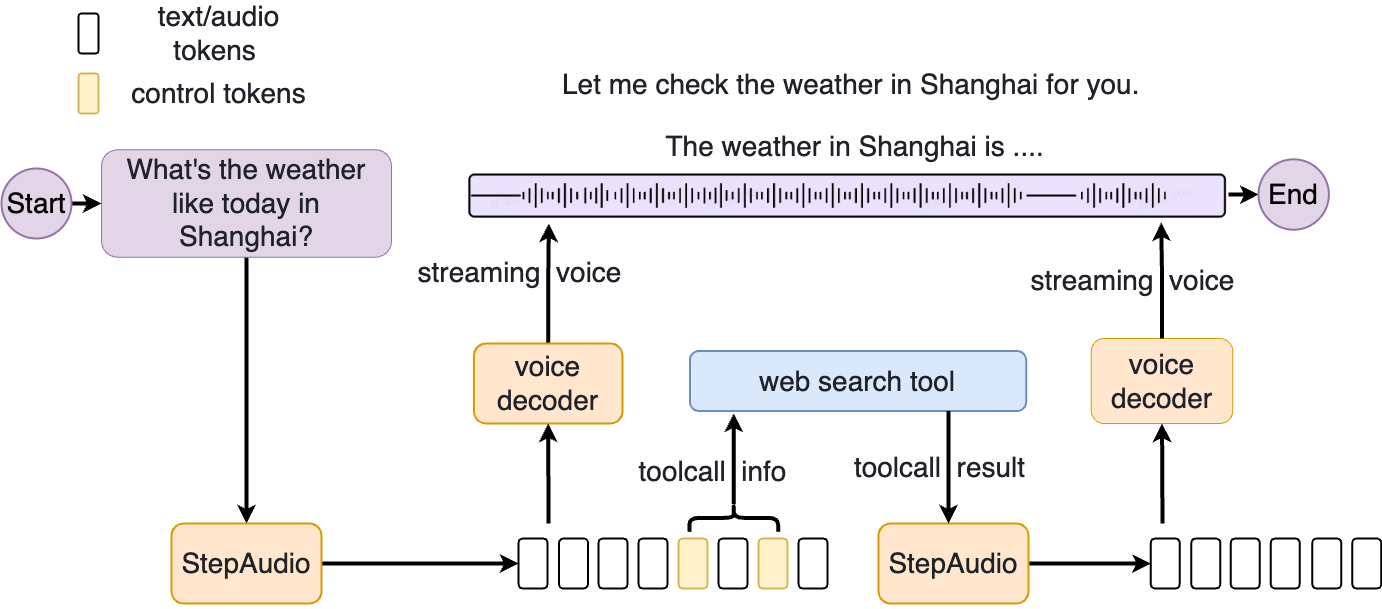}
  \caption{Architecture of asynchronous tool call in Step-Audio. The text processing thread handles tool calls while the audio generation thread produces speech streams concurrently.}
  \label{fig:toolcall}
\end{figure}

\noindent The proposed Step-Audio system enables real-time tool call during voice interactions. Due to the substantial bitrate disparity between real-time text responses and their corresponding audio streams, our framework achieves asynchronous tool invocation while maintaining seamless voice interaction. As illustrated in Figure \ref{fig:toolcall}, the architecture decouples text-based tool processing from audio generation pipelines, allowing parallel execution of external service queries (e.g. knowledge retrieval) and speech synthesis. This design eliminates waiting time for audio rendering when tool calling is required, significantly enhancing interaction fluidity.

\section{Conclusion}
\noindent In this paper, we present Step-Audio, an innovative framework for real-time voice interaction. During the pretraining stage, our dual-codebook speech tokenizer bridges textual and acoustic modalities using 3.3T tokens of multi-modal data, establishing cross-modal alignment. In the post-training phase, we conducted task-specific SFT for TTS and ASR tasks, while implementing SFT with diversified high-quality datasets combined with RLHF for AQTA tasks to enhance response quality, enabling fine-grained control over emotional modulation, dialect adaptation, and prosodic pattern generation. Through engineering innovations including speculative streaming with sliced latency compensation and efficient full-duplex coordination, Step-Audio achieves fluid conversational dynamics. The performance metrics from benchmark evaluations across tasks including ASR, TTS, and AQTA demonstrate Step-Audio's exceptional capabilities in speech dialogue.

\section{Future Work}
\noindent In this work, we have demonstrated Step-Audio's current capabilities in cross-modal integration between speech and text, as part of Step-Audio's initial implementation aimed at trimodal systems. Moving forward, three key areas warrant further exploration: first, extending the framework to achieve native trimodal understanding that incorporates vision, speech, and text; second, enhancing pure voice dialogue efficiency by eliminating intermediate cross-modal conversions in AQAA scenarios; and third, implementing deep-thinking-enhanced tool calls to enhance intelligent interaction capabilities with external knowledge bases.



\bibliographystyle{apacite}    

\bibliography{ref}

\section*{Acknowledgment}
\noindent We designate core contributors as those who have been involved in the development of Step-Audio throughout its entire process, while contributors are those who worked on the early versions or contributed part-time. All contributors are listed in \textbf{alphabetical order by first name}.
\begin{itemize}[leftmargin=*]

\item \textbf{Core Contributors}:
  \begin{itemize}
    \item \textbf{Audio Model and Training:} Ailin Huang, Boyong Wu, Bruce Wang, Chao Yan, Chen Hu, Chengli Feng, Fei Tian, Feiyu Shen, Jingbei Li, Mingrui Chen, Peng Liu, Ruihang Miao, Wang You, Xi Chen, Xuerui Yang, Yechang Huang, Yuxiang Zhang, Zheng Gong, Zixin Zhang.
    \item \textbf{Foundation Model and Training:} Hongyu Zhou, Jianjian Sun.
    \item \textbf{Infrastructure:} Brian Li, Chengting Feng, Changyi Wan, Hanpeng Hu, Jianchang Wu, Jiangjie Zhen, Ranchen Ming, Song Yuan, Xuelin Zhang, Yu Zhou.
    \item  \textbf{Data and Evaluation}: Bingxin Li, Buyun Ma, Hongyuan Wang, Kang An, Wei Ji, Wen Li, Xuan Wen, Xiangwen Kong, Yuankai Ma, Yuanwei Liang, Yun Mou.
  \end{itemize}
\item \textbf{Contributors:}
Bahtiyar Ahmidi, Bin Wang, Bo Li, Changxin Miao, Chen Xu, Chenrun Wang, Dapeng Shi, Deshan Sun, Dingyuan Hu, Dula Sai, Enle Liu, Guanzhe Huang, Gulin Yan, Heng Wang, Haonan Jia, Haoyang Zhang, Jiahao Gong, Junjing Guo, Jiashuai Liu, Jiahong Liu, Jie Feng, Jie Wu, Jiaoren Wu, Jie Yang, Jinguo Wang, Jingyang Zhang, Junzhe Lin, Kaixiang Li, Lei Xia, Li Zhou, Liang Zhao, Longlong Gu, Mei Chen, Menglin Wu, Ming Li, Mingxiao Li, Mingliang Li, Mingyao Liang, Na Wang, Nie Hao, Qiling Wu, Qinyuan Tan, Ran Sun, Shuai Shuai, Shaoliang Pang, Shiliang Yang, Shuli Gao, Shanshan Yuan, Siqi Liu, Shihong Deng, Shilei Jiang, Sitong Liu, Tiancheng Cao, Tianyu Wang, Wenjin Deng, Wuxun Xie, Weipeng Ming, Wenqing He, Wen Sun, Xin Han, Xin Huang, Xiaomin Deng, Xiaojia Liu, Xin Wu, Xu Zhao, Yanan Wei, Yanbo Yu, Yang Cao, Yangguang Li, Yangzhen Ma, Yanming Xu, Yaoyu Wang, Yaqiang Shi, Yilei Wang, Yizhuang Zhou, Yinmin Zhong, Yang Zhang, Yaoben Wei, Yu Luo, Yuanwei Lu, Yuhe Yin, Yuchu Luo, Yuanhao Ding, Yuting Yan, Yaqi Dai, Yuxiang Yang, Zhe Xie, Zheng Ge, Zheng Sun, Zhewei Huang, Zhichao Chang, Zhisheng Guan, Zidong Yang, Zili Zhang.

\item \textbf{Project Sponsors:}
Binxing Jiao, Daxin Jiang, Heung-Yeung Shum, Jiansheng Chen, Jing Li, Shuchang Zhou, Xiangyu Zhang, Xinhao Zhang, Yibo Zhu.
\item \textbf{Corresponding Authors:} Daxin Jiang (\url{djiang@stepfun.com}), Shuchang Zhou (\url{scotzhou@stepfun.com}), Chen Hu (\url{hatcher@stepfun.com}), ~Bruce Wang~(\url{brucewang@stepfun.com})

\end{itemize}

\noindent
\end{document}